\documentclass[runningheads]{llncs}
\usepackage{graphicx}
\usepackage{amssymb}
\usepackage{amsmath}
\begin{document}
\title{Automatic Text Extractive Summarization Based on Graph and Pre-trained Language Model Attention}
%
%
\author{Yuan-Ching  Lin
\and
Jinwen Ma
}
\authorrunning{Y. Lin and J. Ma}
%
\institute{Department of Information and Computational Sciences, School of Mathematical Sciences and LMAM, Peking University, Beijing, 1000871, China\\
\email{\{yuanchinglin,jwma\}@pku.edu.cn}}
\maketitle              
\begin{abstract}
Representing a text as a graph for obtaining automatic text summarization has been investigated for over ten years. With the development of attention or Transformer on natural language processing (NLP), it is possible to make a connection between the graph and attention structure for a text. In this paper, an attention matrix between the sentences of the whole text is adopted as a weighted adjacent matrix of a fully connected graph of the text, which can be produced through the pre-training language model.  The GCN is further applied to the text graph model for classifying each node and finding out the salient sentences from the text. It is demonstrated by the experimental results on two typical datasets that our proposed model can achieve a competitive result in comparison with sate-of-the-art models.

\end{abstract}

\section{Introduction}
As a major task of breakthrough, automatic text summarization has attracted more and more attention recently thanks to the development of deep learning and artificial intelligence. It aims at contracting a piece of text into a summary or an abstract automatically. In general, there are two kinds of methods for automatic text summarization. The first kind of methods are abstractive summarization methods which generate a summary directly from the text. But their results are not so satisfactory, being unstable and unreadable. However, the second kind of methods are extractive summarization methods which can extract some salient sentences from the text so that the results are more stable and readable, usually gain the higher scores on the evaluation index. 

In fact, the extracted or selected sentences through an extractive summarization method ensure that the summary is meaningful and informative. Although the summary may contain certain redundant information, it can represent the key ideas for the whole text. Moreover, an extractive summarization method can be considered as a multi-label classfier, which is more effective on training due to the fewer parameter and relatively simple structure.

Currently, Natural Language Processing (NLP) with neural network architecture has become the mainstream, with several outstanding results \cite{see2017get,Cheng2016,nallapati2016summarunner}. Especially, the Transformer\cite{vaswani2017attention} and attention mechanism \cite{bahdanau2014neural} play extraordinary performance to extract the features within the text, making a breakthrough in all evaluation indexes on various NLP tasks. Such an approach has been already adapted in text summarization task, and the experiments demonstrate that attention mechanism can locate the text with rich semantic information. Moreover, the pre-training language model based on attention mechanism
\cite{radford2019language,devlin2018bert,liu2019roberta} can collect the text information in a mode of unsupervised learning, with properly fine-tuning training, we can use limited computation resources to reach high-quality results.

In this paper,  we try to analyze the text summarization task with an perspective of text structure through a graph model. Inspired by the development and application of Graph Convolutional Network (GCN)\cite{kipf2016semi}, we represent a piece of text as a graph and select the key sentences by the message passing and node embedding in GCN. The cross-attention among the sentences in a text can be regarded as a structural data , and we can discover such a feature in the attention-based pre-trained language model.

We then investigate the connection between the attention and the graph. An attention matrix between the sentences is constructed as a weighted adjacent matrix of a fully connected graph. In our text graph model, each node represents a sentence. The node feature and the adjacent matrix can be obtained through sentence embedding and attention layers in the pre-trained transformer, respectively. We further apply the GCN to the text graph model for classifying each node and finding out the salient sentences from the text.

The experimental results demonstrate that our proposed model can achieve a comparable result on two datasets-CNN/dailymail and Nessroom with the ROUGE\cite{lin2004rouge} index.  What's more, we can obtain such results with only about 5 million parameters, 1 percent compared to the sate-of-the-art model, and an accepted loss in precision. As a result, our approach of extractive text summarization is more effective.

\section{Related Work}

Graph-based summarization model has been used in several previously published studies. The early works focus on the variant of PageRank\cite{page1999pagerank} algorithm, which assumed that important sentences would connect to other important ones. Textrank\cite{mihalcea2004textrank} and Lexrank\cite{Erkan_2004} adopted such concept, they defined the relation between sentences by tf-idf\cite{jones1972statistical} similarity between sentences and build the text graph, then ranking all the sentences through graph iterating and return the first few sentences as the summaries.

In current work, the neural network architecture has joined with the graph method thanks to the improvement of sentence representation and graph neural network. \cite{yasunaga2017graph} applied the GCN model to find the salient sentences. They use three different rules (including sentence similarity) to build the text graph and encode the sentences to a fix-dimensional vector through recurrent network networks (RNN).

However, building the sentence relation by rules may be limited to semantic information. The statistic method such as tf-idf or discourse relations\cite{christensen2013towards} just reflect part of the information, but it is insensitive to the word meaning like the case of synonyms or antonyms. We improved this issue by constructing the text graph based on learning. The attention matrix generated by the language model contains much more information, and it is also more flexible in representing the sentence relations in different contexts.

\section{Attention Weigh Matrix Text Graph Model}
For a given text, we first input it into the pre-trained transformer model and obtain the sentence embedding and the attention matrix. After modified the attention matrix to represent the sentence relation matrix, we set a threshold to binarize the matrix and it can be viewed as the adjacent matrix for the GCN model. By the message passing between each sentence node, the model would output the score for each sentence node, and we can select the highest few sentences as the summary for the text.

\subsection{Sentence embedding and graph building}

With the assumption that essential sentences in a text should receive more attention from other sentences, we can infer that the GCN model will capture the essential information from the relation network structure of the sentences.

To understand the relative between each sentence $s_i$ in a text $d = (s_1, ..., s_n)$, we input all the sentences to the pre-trained BERT model and modified the input layer as the BertSum\cite{liu2019fine} model. First, an article is split into several sentences, each sentence is tokenized, then a classification token ([CLS]) is added before the sentence and [SEP] is added at the end of the sentence as a separator, and finally the whole article is ended with a separation token([SEP]). In the segmentation embedding part, the 1 and 0 tokens are used to interactively appear at the position of each article. The last is a point-to-point summation of the sublayer of the three inputs, and the vector of each [CLS] position can be taken as the representative vector of the sentences.

\begin{figure}[h]
    \centering
    \includegraphics[scale=0.3]{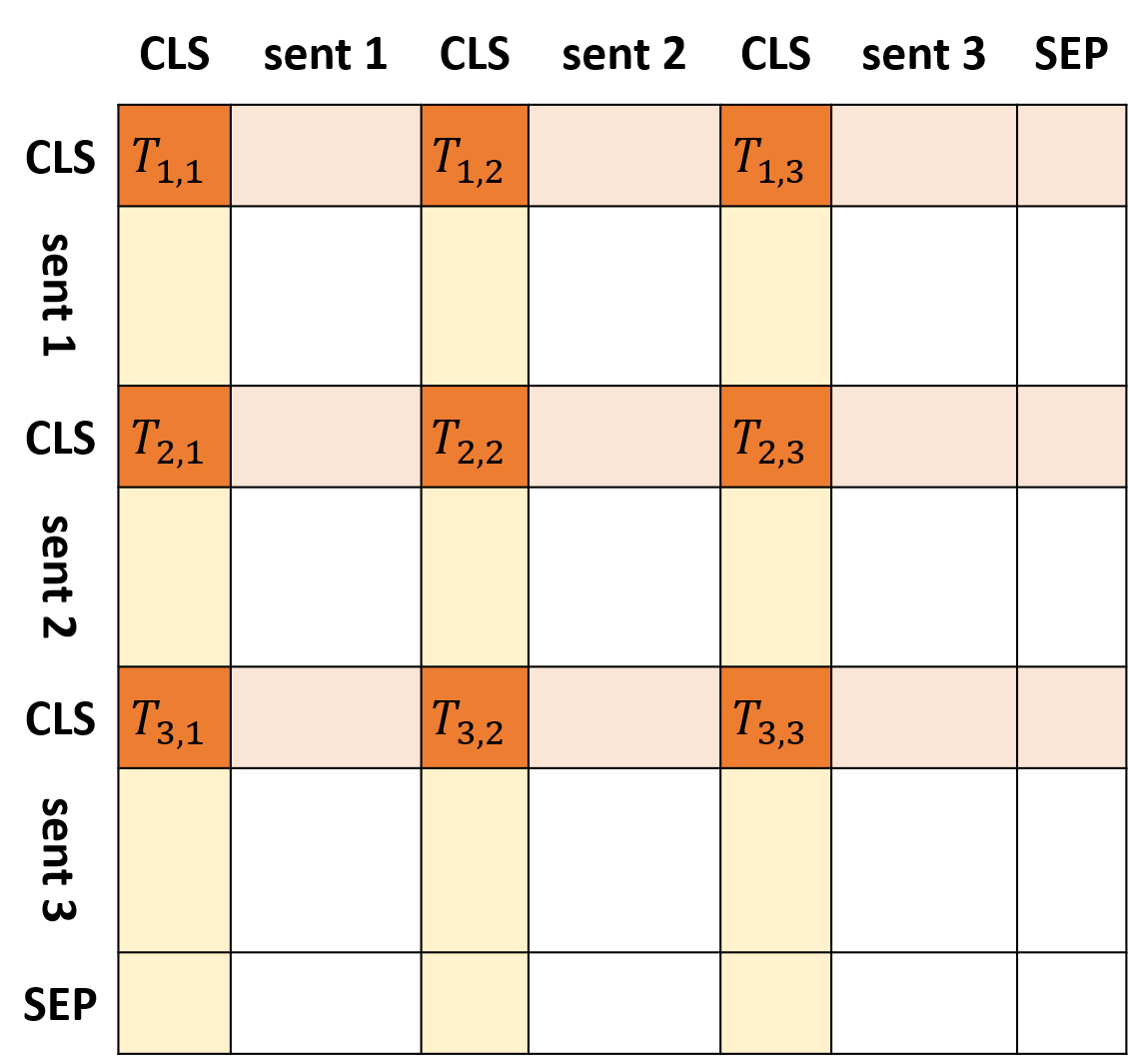}
    \caption{Building the inter-sentences attention matrix from BERT attention matrix. If the original text contains 3 sentences, select the [CLS] token from each row and column to form a 3×3 matrix}
    \label{sentattn}
\end{figure}

The encoding layer of BERT contains a large amount of information about the attention mechanism, and the attention matrix can be used independently. In the process of self-attention mechanism in each layer of BERT, an attention matrix of all tokens interactions is generated. By picking out the [CLS] position representing each sentence in each attention matrix, an inter-sentence attention matrix can be obtained. Figure \ref{sentattn} shows the method of generating the inter-sentences attention matrix. Taking out the Attention of each pair of sentences is not a true inter-sentence self-attention, because the original attention is a probability distribution for all the dimensions of the BERT (512), to make each sentence form a probability distribution of attention to the other sentences, it needs to be fed into softmax function again.

\begin{figure}[h]
    \centering
    \includegraphics[scale=0.5]{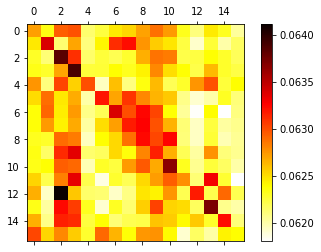}
    \includegraphics[scale=0.5]{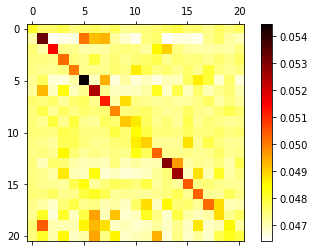}
    \caption{The two image represent the same text, obtained by different attention heads. Each head will focus on different information and generate different attention result. Adding more information into the model helps to improve the semantic richness. }
    \label{att_res}
\end{figure}

\begin{figure}[h]
    \centering
    \includegraphics[scale=0.5]{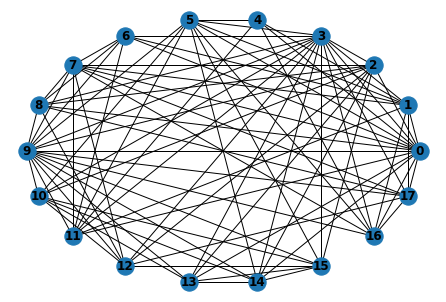}
    \caption{Adjacent matrix of text structure}
    \label{adj-graph}
\end{figure}

Figure \ref{att_res} shows the visualized inter-sentence attention matrix of a text data. Observation is that each sentence has the most attention to itself and various attention to other sentences. We find that the sentence attention matrix has captured some basic semantic information and that can be used to represent the structured text. After setting a threshold and binaries the attention value as 0 and 1, it can be considered as the adjacent matrix in the GCN model. Note that since attention is formed from sentence to sentence in the attention matrix, the resulting graph is a directed graph, but an undirected graph must be used in the graph convolution network, so all directed edges are converted to undirected edges, and thus the directed graph is converted to an undirected graph. Figure \ref{adj-graph} shows the connection of a graph for a given text. It is worth noting that parameter training is not needed during the process of producing the adjacent matrix, which helps to stabilize the GCN model and reduce the computational.

\subsection{Graph convolutional network}

In a text, the salient sentences that represent the summary of the text are usually associated with more sentences, and these associated sentences are also important supports for the theme of the whole text and have a certain position in the text. By means of information transfer in GCN, information can be aggregated into summary sentences to find out the summary of a text. Therefore, We can train the GCN model as a classifier to assign the $\{1,0\}$ label for each sentence, which shows whether a sentence is the summary in text.

The inputs of the GCN model are the sentence node vectors and the adjacent matrix. Let $X \in \mathbb{R}^{N \times d}$ be the vectors of N sentence and $A \in [0,1]^{N \times N}$ be the adjacent matrix, where d is the dimensions for each vector. The process of GCN can be represent as
\begin{equation}
GCN(X, \tilde{A}) = relu(\tilde{A}\cdot FC(X)),
\end{equation}
where $\tilde{A}$ is the normalized adjacent matrix\cite{kipf2016semi}, $FC$ is fully connected neural network. We design a readout function, which combine the information from each GCN model, as output based on the structure of MPNN\cite{gilmer2017neural}. Define $H^2$ as the node embedding processed by two different GCN layer of the input $H^0 = X$, and $FC1$, $FC2$ as 2 independent fully connected layer. The readout function can be computed as
\begin{equation}
R=\sigma\left(FC_{1}\left(H_{i}^{2} \mid H_{i}^{0}\right)\right) \odot FC_{2}\left(H_{i}^{2}\right),
\end{equation}
where $\odot$ represent the  point-to-point product and $\sigma$ is the sigmoid function. Now the output $R$ is the updated node vector, through a fully connected MLP layer we can condense the information into one dimension to obtain the prediction
\begin{equation}
\hat{y} = MPL(R),
\end{equation}
where $\hat{y}=P(y|D)$, the prediction vector of a given text.

In the BERT-base model, there are 12 layers and each layer has 12 different attention heads. We choose the output attention of the first layer to produce our adjacent matrix because it remains the most information from the original text data. To merge the adjacent information of all attention heads, we modified the node updating formula as
\begin{equation}
\begin{aligned}
H^1 &= concat(GCN1(X,\tilde{A}_1), GCN1(X,\tilde{A}_2), \dots, GCN1(X,\tilde{A}_{12})), H^1 \in \mathbb{R}^{N \times d_1 \times 12} \\
H^2 &= \text{concat}(GCN2(H^1,\tilde{A}_1), GCN2(H^1,\tilde{A}_2), \dots, GCN2(H^1,\tilde{A}_{12})), H^2 \in \mathbb{R}^{N \times d_2 \times 12},
\end{aligned}
\end{equation}

where each GCN share the parameters, $d_1$ and $d_2$ are the output dimension of GCN. We can compute $H^1$ and $H^2$ as above to get $\hat{y}$. Therefore, the attention matrix is extracted from language model as the semantic information to evaluate the importance of sentences.

\section{Experimental Results}

\subsection{Datasets}
We test our framework on 2 common datasets in summarization task to verify the feasibility. For all datasets, we use greedy algorithm to label the summaries in the text, where we pick N sentences with the highest ROUGE-2 score and N is the average number of summaries of each dataset. Additionally, we conduct two experiments for each dataset, one is for the original dataset, another is for the modified dataset with selected samples. Due to the widely various text length, we modified each dataset by removing samples with extreme lengths. 

\subsubsection{CNN/Dailymail} is a summarization dataset provided by Hermann et al.\cite{hermann2015teaching} with long articles and each text contains several summaries. The texts mainly are news data, and the summaries are writted by specialists. There are 287,226 pairs for training and 11,490 pairs for testing. The average length of a text is about 30 sentences, and each text contains about 3.72 summary sentences.

\subsubsection{Newsroom} is collected from website texts such as social media or news\cite{grusky2018newsroom}. The feature of the dataset is that it covers texts from a variety of different pipelines, and the text length varies widely. In total, the dataset contains about 1.3 million data pairs with an average text length of 658 words and about 15 sentences, with each text containing an average of 1.33 summary sentences.

\subsection{Implementation Detail}
In each dataset task, we choose K sentences with the highest output score as the summaries, where K is the average number of summaries per text. When conducting the  extra experiment, we train and test on the modified dataset. We use BERT-base as the language model to obtain the attention matrix. All the experiments in this paper are conducted with Tensorflow2.3.0 framework and NVIDIA GeForce GTX 1080, 8GB as GPU.
 
\subsection{Summarization Perforamnce and Comparison}
We compare our method with the mainstream summarization work. Table \ref{cnn_ex} shows the results on CNN/Dailymail, where Lead-3 is the baseline, which simply chooses 3 sentences at the beginning of the text as summaries. The upper block is the testing result of our model, the lower ones are selected methods in recent years. The model followed by * indicates the experiment conducted with the modified dataset. It can be observed that our model surpasses the baseline model and achieves a competitive result among other works.

\begin{table}[h]
  \centering
  \begin{tabular}{lccc}
  method               & ROUGE-1 & ROUGE-2 & ROUGE-L \\ \hline
  GCN\_Attn            & 29.85   & 10.76   & 23.85   \\
  GCN\_Attn*  & 26.64 & 8.37 & 21.22 \\\hline
  Lead-3 & 22.22 & 8.32 & 21.17 \\
  GPT-2\cite{radford2019language}         & 29.34   & 8.27    & 26.58   \\
  C2F-ALTERNATE\cite{ling2017coarse} & 31.1    & 15.4    & 28.8    \\
  PEGASUS\cite{zhang2020pegasus} & 44.17 & 21.47& 41.11 \\\hline 
  \end{tabular}
  \caption{Result comparison on CNN/Dailymail dataset, the number represents the F1 score percentage of each measure}
  \label{cnn_ex}
\end{table}

In table \ref{nr}, we saw a similar result on Newsroom dataset. It is worth mentioning that there is a dramatic improvement in the result of modified dataset. The potential reason is that the extreme variety of Newsroom dataset will cause unstable training result hence lower the ability of model to label correct summaries, so when we select the text with moderate length will help for training result.  

\begin{table}[h]
  \centering
  \begin{tabular}{lccc}
      method & ROUGE-1 & ROUGE-2 & ROUGE-L \\\hline
      GCN\_Attn & 27.42 & 19.54 & 26.65 \\
      GCN\_Attn* & 36.75 & 29.39 & 36.29 \\
      \hline
      Lead-2 & 28.43 & 20.57 & 28.66 \\
      Lead-2* & 44.28 & 36.97 & 44.98 \\
      Pointer-N & 26.02 & 13.25 & 22.43 \\
      Pointer-N* & 39.11 & 27.95 & 36.17 \\ PEGASUS\cite{zhang2020pegasus} & 45.15 & 33.51 & 41.33\\\hline
  \end{tabular}
  \caption{Results on Newsroom dataset, where Point-N is the dataset\cite{grusky2018newsroom} author's training using pointer-generator\cite{see2017get}, and Point-N* is the same method training with selected dataset.}
  \label{nr}
\end{table}

\subsection{Discussions}
Observing the experiment results of the two datasets, we can infer that in the short text, filtering the dataset can improve the ability of the model to select the summaries, while for the long text it shows little help. The reason may come from the limitation of the Bert-base model, which can only input up to 512 tokens, and it is insufficient for long text, so that even we removing the extreme short or long text, it still performs unstable.

In the model proposed in this paper, although the evaluation indexes are lower than those of the state-of-the-art method, they can reach a level comparable to the existing methods and exceed the baseline model Lead-N, which proves that the algorithm idea of this paper is feasible. In addition, the method of this paper has the characteristics of fast training and few model parameters, which is an advantage compared with other models. For example, PEGASUS is the best model in all the datasets, but the number of parameters is as high as 568 million, while the GCN model proposed in this paper is only 504839, which are much smaller than PEGASUS and can be executed faster in the training and inference process. In the absence of computational memory and limited time, the model in this paper is of more practical value.

\section{Conclusion}

We have proposed a new idea to construct the adjacent matrix for GCN in text summarization task. No additional training is required for the language model to generate the attention matrix and it has already contained the semantic information. We represent the text as a graph model with nodes as sentences, and find the important sentences as summaries by message passing.

Actually, we design the algorithm to explore new ideas to solve the text summarization problem and successfully achieve results that can be compared with the mainstream methods in recent years, even with the advantage of fewer parameters and faster training in terms of efficiency. There are still some aspects of the relationship between graphs and language models that can be explored, and we will continue our experiments in an attempt to improve the accuracy and usefulness of graph summarization models.

\section{Acknowledgment}
This work was supported by the National Key Researchand Development Program of China under grant 2018AAA0100205.

\bibliographystyle{splncs04}
\bibliography{refer}
\end{document}